\documentclass[sigconf, 10pt]{acmart}

\usepackage{caption}
\usepackage{subcaption}
\usepackage{graphicx}
\usepackage{placeins}
\usepackage{xspace}
\usepackage{colortbl}
\usepackage{cleveref}
\usepackage{lipsum}
\usepackage[normalem]{ulem}
\usepackage{enumitem}
\usepackage{geometry}
\usepackage{amsmath}
\setlist[itemize]{noitemsep, topsep=0pt}
\usepackage[utf8]{inputenc}
\usepackage[T1]{fontenc}
\usepackage{adjustbox}
\usepackage{multirow}
\usepackage[normalem]{ulem}
\newcommand\sbullet[1][.5]{\mathbin{\vcenter{\hbox{\scalebox{#1}{$\bullet$}}}}}
\usepackage[most]{tcolorbox}
\usepackage{capt-of}

\newcommand{\ts}{\textsuperscript}

\usepackage{soul}
\usepackage{hyperref}
\hypersetup{
    colorlinks=true,
    linkcolor=blue,
    filecolor=magenta,
    urlcolor=cyan,
}

\usepackage{siunitx}
\usepackage{tikz}
\newcommand*\circled[1]{\tikz[baseline=(char.base)]{
            \node[shape=circle,draw,inner sep=0.8pt] (char) {#1};}}
\newcommand*\circledd[1]{\tikz[baseline=(char.base)]{
            \node[shape=circle,draw,inner sep=0.7pt] (char) {#1};}}

\makeatletter
\DeclareRobustCommand\onedot{\futurelet\@let@token\@onedot}
\def\@onedot{\ifx\@let@token.\else.\null\fi\xspace}

\def\eg{\emph{e.g}\onedot} \def\Eg{\emph{E.g}\onedot}
\def\ie{\emph{i.e}\onedot} \def\Ie{\emph{I.e}\onedot}
\def\cf{\emph{c.f}\onedot} \def\Cf{\emph{C.f}\onedot}
\def\etc{\emph{etc}\onedot} \def\vs{\emph{vs}\onedot}
\def\wrt{w.r.t\onedot} \def\dof{d.o.f\onedot}
\def\etal{\emph{et al}\onedot}
\makeatother

\AtBeginDocument{%
  \providecommand\BibTeX{{%
    \normalfont B\kern-0.5em{\scshape i\kern-0.25em b}\kern-0.8em\TeX}}}

\newcommand{\kc}[1]{{\color{orange}[Kichang: #1]}}

\newcommand{\jk}[1]{{\color{blue}[Jeonggil: #1]}}

\newcommand{\urlcolor}[1]{\textcolor{magenta}{#1}}
\newcommand{\system}{MetaSE}

\begin{document}
\title[]{Metacognitive Selective Ensemble for Mobile Systems}



\author{Sungmin Lee}
\email{i.am.sungmin.lee@yonsei.ac.kr}
\affiliation{%
  \institution{Yonsei University}
  \country{}
}

\author{Kichang Lee}
\email{kichang.lee@yonsei.ac.kr}
\affiliation{%
  \institution{Yonsei University}
  \country{}
}

\author{Joonhee Lee}
\email{neo81389@yonsei.ac.kr}
\affiliation{%
  \institution{Yonsei University}
  \country{}
}

\author{JaeYeon Park}
\email{jaeyeon.park@dankook.ac.kr}
\affiliation{%
  \institution{Dankook University}
  \country{}
}

\author{Songkuk Kim}
\email{songkuk@yonsei.ac.kr}
\affiliation{%
  \institution{Yonsei University}
  \country{}
}

\author{JeongGil Ko}
\email{jeonggil.ko@yonsei.ac.kr}
\affiliation{%
  \institution{Yonsei University}
  \country{}
}


\begin{abstract}
Deep ensembles improve robustness in mobile sensing, but repeatedly executing many models over continuous sensor streams is costly. Selecting only a few members reduces this cost, yet adaptive selection often requires additional model execution to obtain reliable evidence about inactive candidates. We present \system{}, an active ensemble framework that exploits short-term persistence in per-model reliability. \system{} maintains a small active set across windows, uses post-execution evidence to reject unreliable members, and invokes lightweight routing only when replacement is needed. This stateful design accesses the diversity of a larger pool without repeated full-pool evaluation. Across four HAR datasets and four model architectures, \system{} consistently improves over a fixed three-model ensemble and achieves accuracy comparable to substantially more expensive adaptive and full-ensemble inference. On a Raspberry Pi 4B, \system{} is 2.7$\times$ faster and uses 69\% less memory than full ten-model inference.
\end{abstract}
\settopmatter{printfolios=false} 
\settopmatter{printacmref=false} 
\renewcommand\footnotetextcopyrightpermission[1]{} 
\maketitle

\section{Introduction}
\label{sec}

Modern mobile sensing applications increasingly rely on deep learning models that operate continuously across different users, devices, sensor placements, and environments~\cite{lane2015can,lane2017squeezing,yao2018deep,wang2019deep,zhang2022deep}. Such deployment heterogeneity makes robust inference challenging since a model that performs well under one condition can degrade under another~\cite{stisen2015smart,ovadia2019can,mathur2019mic2mic,kwapisz2011activity,reiss2012introducing}. Ensemble methods, which combine the predictions of multiple models, provide a practical way to improve robustness by leveraging models that make different errors~\cite{dietterich2000ensemble,lakshminarayanan2017simple,fort2019deep,sagi2018ensemble,breiman1996bagging,freund1997decision,breiman2001random}. This property is particularly useful in mobile sensing, where different models can complement one another across users and sensing conditions.

Yet conventional ensemble prediction does not necessarily realize all of the diversity available across its members. As more diverse members are added, the fraction of inputs for which at least one member is correct, which we refer to as \textit{oracle coverage}, can remain higher than the accuracy produced by their final aggregation~\cite{liu1999ensemble,brown2003ambiguity,wood2023unified}. Voting or probability averaging must make a final prediction without knowing which member is correct, so their effectiveness depends on how correct predictions and errors are distributed across members~\cite{tumer1996error, kuncheva2003measures}. A correct minority may not determine the final prediction, while correlated failures can make additional votes redundant. The gap between ensemble prediction and oracle coverage, therefore, represents useful model diversity that conventional aggregation does not fully exploit.

However, simply executing more members to access this diversity is unattractive on a mobile device. Each additional model increases inference computation and energy consumption and can also increase latency~\cite{han2016deep,howard2017mobilenets,sandler2018mobilenetv2}. These costs accumulate continuously in sensing applications that process window after window throughout deployment. A more attractive approach is to keep only a small \textit{active set} from a larger model pool and change its members when different models become useful. This preserves access to the larger pool without requiring full-ensemble inference at every sensing window.

Prior work reduces ensemble inference costs by executing only a subset of member models, selected either statically before deployment or adaptively for each input~\cite{zhou2002ensembling,caruana2004ensemble,shazeer2017outrageously,wang2018skipnet,britto2014dynamic,cruz2018dynamic}. Static selection incurs little runtime overhead but cannot account for input-dependent changes in model usefulness. Adaptive selection offers greater flexibility, but reliable decisions are difficult without runtime labels and may require additional model execution. On resource-constrained devices, such extra inference and frequent model replacement can offset the intended latency and energy savings~\cite{fang2018nestdnn}. The key challenge is therefore to make reliable per-model selection decisions while minimizing both inference and replacement overhead.

To address this problem, we exploit the temporal continuity of sensor streams. Nearby windows often share similar sensing conditions, allowing per-model reliability to persist over short intervals~\cite{sponner2023temporal,xu2018deepcache}. This continuity enables active members to provide low-cost runtime evidence, as their predictions and intermediate outputs are already available from normal inference, whereas evaluating inactive members requires additional forward passes. Based on this asymmetry, we use rich post-execution evidence to retain or remove active members and lightweight pre-execution information only to select their replacements. Otherwise, the active set is carried forward across windows, avoiding repeated pool-wide evaluation and reducing both inference and model-replacement overhead.

Building on this strategy, we present \system{}, an active ensemble framework for continuous mobile sensing. After a one-time full-pool initialization, \system{} maintains a small active set across consecutive sensing windows. Each active member is checked during normal inference, while the active set itself is kept by default. When replacement is triggered, \system{} follows a \textit{reject-then-route} procedure. It first identifies the active member that should no longer be used and then selects an alternative from the inactive model pool using lightweight pre-execution information. Unaffected active members are preserved, and the replacement participates from the following window. Thus, model routing is invoked only when replacement is needed rather than reconstructing the selected ensemble at every window.

Implementing this strategy first requires reliable member rejection without runtime labels. Self-confidence is readily available but can remain high even for incorrect predictions, particularly under distribution shift~\cite{ovadia2019can}. \system{} therefore attaches a lightweight rejecter to each member, combining its prediction and feature representation with evidence from the other active members and short-term prediction stability. Because these signals are already available from active-set inference, member reliability can be evaluated without executing inactive models.

Once replacement is triggered, a second challenge is selecting a useful inactive member with minimal runtime overhead. Probing candidates would provide stronger evidence, but every additional forward reduces the savings of selective execution. \system{} instead uses a lightweight \textit{class-conditional routing table} constructed before deployment. Indexed by the rejected member and its predicted class, the table records which other members tend to correct that member's corresponding errors. A replacement can therefore be selected through a simple lookup without executing inactive candidates, and routing occurs only when the active set requires an update.

We evaluate \system{} across four HAR datasets and four model architectures
against fixed, full-ensemble, and adaptive-selection baselines, together with
an embedded-device deployment.
\system{} consistently improves over a fixed three-model ensemble across most
configurations and achieves accuracy comparable to full ten-model inference
while executing only three members during regular inference.
Under the same three-model execution budget, it also consistently
outperforms representative dynamic ensemble selection methods.
On a Raspberry Pi 4B, \system{} is 2.7$\times$ faster and uses 69\% less
memory than full ten-model inference, showing that the diversity of a larger
pool can be exploited without repeatedly executing the entire ensemble.

We make the following contributions.

\noindent{}\textbf{$\bullet$} We characterize short-term persistence in per-model reliability over continuous sensor streams. Member reliability remains locally correlated across nearby inference windows, making current runtime behavior useful for subsequent active-set decisions.

\noindent{}\textbf{$\bullet$} We formulate adaptive ensemble execution around the tradeoff between runtime evidence and systems overhead. Rich model-specific evidence is available after an active member has already been executed, whereas obtaining comparable evidence from inactive members requires additional model forwards. Changing the active set can further introduce replacement overhead. This motivates asymmetric use of runtime information together with stateful active-set maintenance.

\noindent{}\textbf{$\bullet$} We design \system{}, an active ensemble framework that realizes this operating point through reject-then-route execution. \system{} combines lightweight per-member rejection using post-execution evidence with class-conditional routing for replacement selection, while preserving unaffected active members across updates.

\noindent{}\textbf{$\bullet$} We evaluate \system{} across multiple mobile human activity recognition datasets and deployment conditions, jointly examining prediction quality, model-forward cost, and replacement overhead against static, full-ensemble, and adaptive selection baselines.
\section{Background and Related Work}
\label{sec:relwork}
This section reviews prior work on ensemble learning and efficient inference that is most relevant to our problem setting. We first summarize how model diversity is exploited in conventional and selective ensembles, and then discuss adaptive and resource-aware inference methods that motivate deployment-time model selection on mobile devices.
\subsection{Deep Ensembles and Ensemble Selection}

Ensemble methods improve predictive performance by combining models with complementary errors rather than relying on a single predictor. In deep learning, a common form is the \emph{deep ensemble}, where multiple independently trained models are combined at inference time through averaging, voting, or related aggregation rules~\cite{lakshminarayanan2017simple,fort2019deep,sagi2018ensemble}. Such diversity can improve accuracy, robustness, and uncertainty estimation, but conventional deep ensembles typically execute all members for every input.

Ensemble selection reduces this cost or improves prediction by using only part of the available model pool. Static ensemble pruning selects a subset before deployment according to validation performance or ensemble utility~\cite{zhou2002ensembling,caruana2004ensemble}. Because the selected subset is fixed, these methods add little runtime overhead but cannot respond when the relative usefulness of members changes across inputs. This limitation has motivated dynamic forms of classifier and ensemble selection.

\subsection{Dynamic Classifier and Ensemble Selection}

Dynamic Selection selects classifiers on a per-query basis according to their estimated \emph{competence}, which measures how likely each classifier is to correctly classify the current input~\cite{britto2014dynamic,cruz2018dynamic}. Dynamic Classifier Selection chooses a single classifier, while Dynamic Ensemble Selection chooses a subset from the classifier pool. Existing methods differ mainly in how they estimate competence, including local performance around the query, classifier rankings, and probabilistic or learned competence models~\cite{britto2014dynamic,cruz2018dynamic}. This formulation is closely related to our goal because it directly treats member usefulness as input-dependent rather than fixing one ensemble for the entire deployment.

Other methods estimate reliability from the classifiers' current outputs. Confidence-based ensemble selection, for example, selects or weights classifiers based on their prediction confidence, potentially together with classifier-specific credibility information estimated from training data~\cite{nguyen2020ensemble,li2021prediction}. Such output-based evidence reflects how a specific classifier actually responds to the current input, but it is available only after that classifier has been executed. This distinction becomes important when inference cost limits how many members can be evaluated at runtime.

A complementary direction is to select computation before model execution. Learned routing and conditional-computation methods map the current input to a subset of computation, with Mixture-of-Experts (MoE) being a prominent example~\cite{jacobs1991adaptive,shazeer2017outrageously,lepikhin2021gshard,fedus2022switch}. Sparse MoE typically uses a learned router to activate only a small subset of experts within a larger model, thereby avoiding execution of every expert. Our setting instead begins with a pool of full-task ensemble members and focuses on deployment-time management of which members remain active. Accordingly, \system{} uses output-side evidence only for members that are already active, and invokes lightweight routing only when a rejected member must be replaced. This preserves the competence-based view of dynamic selection while avoiding the need to obtain current outputs from every candidate.

\subsection{Efficient Ensemble Inference on Resource-Constrained Devices}

A broad literature reduces deep-inference cost through model compression, pruning, early exiting, cascades, and other forms of adaptive computation~\cite{cheng2018recent,han2016deep,teerapittayanon2016branchynet,bolukbasi2017adaptive,huang2018multi,kaya2019shallow,han2022dynamic,veit2018convolutional}. These approaches commonly vary the amount of computation used for each input. Closely related work has applied the same principle directly to ensembles. Confidence-based adaptive ensembling stops evaluating additional members once the accumulated prediction is sufficiently confident~\cite{inoue2019adaptive}, while learned sequential ensemble methods determine whether another model should be executed for a difficult input~\cite{li2023towards}. These methods can substantially reduce average ensemble computation, but primarily adapt \emph{how much} ensemble inference is performed for each input. \system{} instead keeps a small active set and adapts \emph{which} independently trained members occupy that set over a continuous sensing stream.

Efficient ensemble or ensemble-inspired models have also been studied specifically for sensor-based human activity recognition. Mekruksavanich et al.~\cite{mekruksavanich2024ensemble} propose an ensemble deep network based on CNN--LSTM models for IMU activity recognition. Huang et al.~\cite{huang2022deep} take a different approach with filter activation, using information from multiple networks during training but retaining only a single network for inference. These methods improve HAR accuracy or inference efficiency through the training architecture itself. In contrast, \system{} assumes an available model pool and addresses the runtime problem of selectively executing and replacing its members.

These directions collectively reduce the cost of either model inference or ensemble execution, but continuous mobile sensing adds a systems dimension to adaptive model selection. Runtime evidence is richer for models that have already been executed, while obtaining the same evidence from inactive members requires additional forwards. Maintaining a changing model set can also incur loading and memory-transfer costs on resource-constrained devices. \system{} targets this deployment-time tradeoff by combining post-execution member rejection, replacement-triggered routing, and persistent active-set maintenance.
\section{Motivating Analysis}
\label{sec:feasibility}

We next examine the empirical properties that motivate the design of
\system{}. Our analysis asks three questions. First, does a larger model
pool provide useful alternatives even when only a small subset can be
executed at each sensing window? Second, does per-model reliability
persist long enough to maintain the selected models across nearby
windows rather than repeatedly replacing them? Finally, when an active
model becomes unreliable, can it be identified and replaced without
executing additional candidate models?

\subsection{Opportunity for Selective Ensemble Execution}
\label{sec:feasibility:opportunity}

We first examine whether maintaining a larger model pool is useful when
runtime execution is limited to only a few members. We denote the total
number of models in the pool by $N$ and the number executed at each
sensing window by $K$, where $K < N$. For this motivating study, we use
HHAR~\cite{stisen2015smart}, which contains IMU measurements for six
human activities, and train a pool of $N=10$ Transformer models with the
same training configuration~\cite{vaswani2017attention,wen2022transformers}
but different random initializations. The models therefore have
comparable individual capacity but make different prediction errors.

\begin{figure}[t]
    \centering
    \includegraphics[width=.95\linewidth]{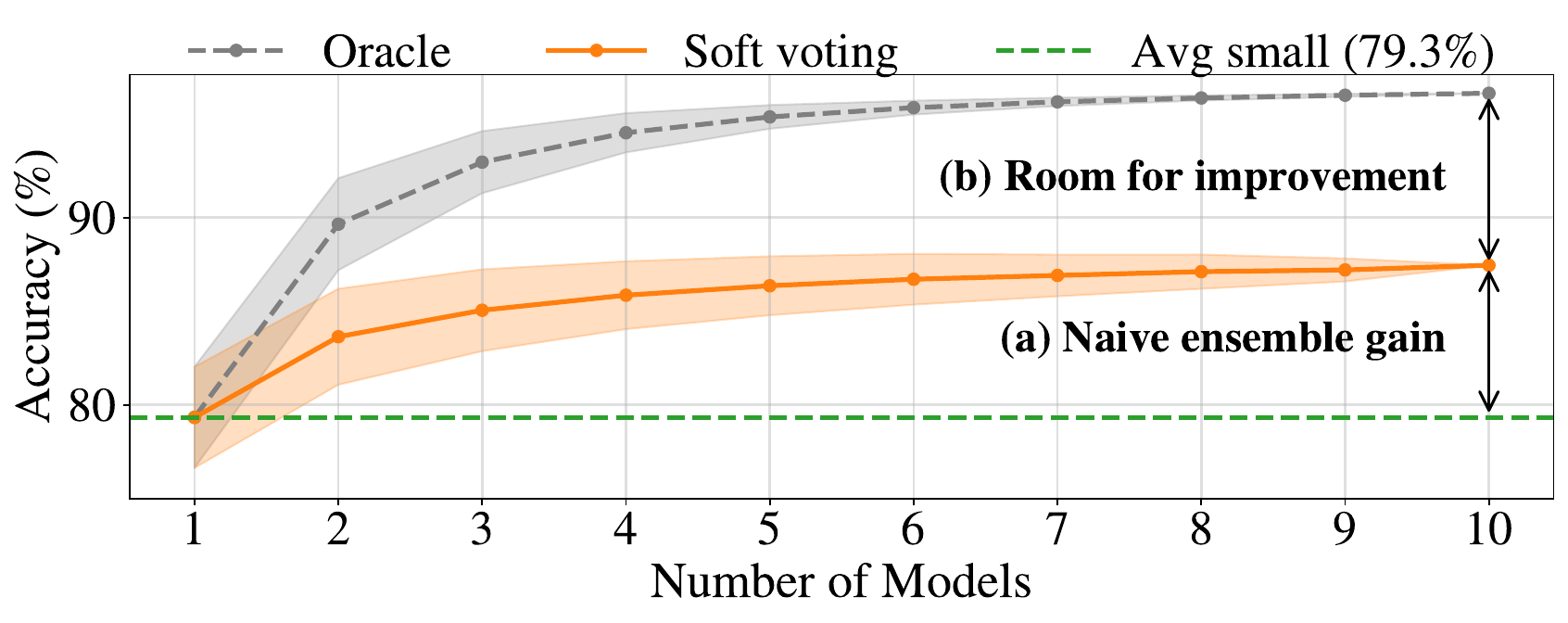}
     \vspace{-2ex}
    \caption{Accuracy as the number of available ensemble members
    increases. The shaded region reports variation across model
    subsets of the same size.}
    \vspace{-3ex}
    \label{fig:motivation_acc_vs_n}
\end{figure}

Figure~\ref{fig:motivation_acc_vs_n} examines how the predictive value
of the pool changes as more models become available. The horizontal
axis represents the number of available models and the vertical axis
reports classification accuracy. Soft voting averages the predictions
of the selected members, whereas the oracle counts an input as correct
if at least one available member predicts it correctly. The shaded
region reports variation across model subsets of the same size.

Soft-voting accuracy improves rapidly with the first few models but
then saturates, whereas oracle coverage continues to increase. Thus,
additional pool members continue to provide complementary correct
predictions even after aggregating more models yields little further
gain. The oracle is used here to expose this complementary diversity,
rather than as the target performance of selective execution. The
result indicates that a larger pool provides useful alternatives from
which a $K$-member ensemble can be selected.

\begin{figure}[t]
    \centering
    \includegraphics[width=\linewidth]{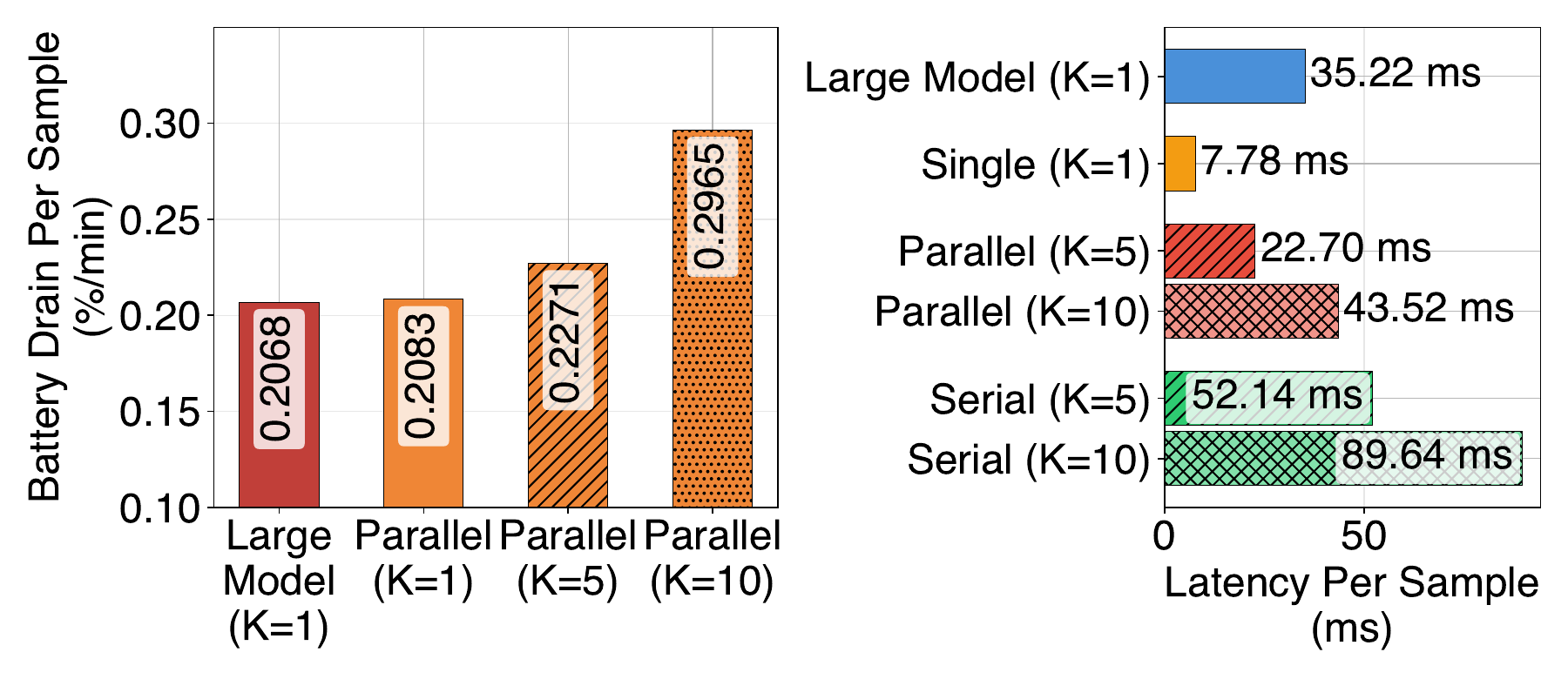}
     \vspace{-5ex}
    \caption{Mobile systems cost as more ensemble members participate
    in inference. Left: battery drain per sample. Right: inference
    latency under parallel and serial execution. Lower is better in
    both plots.}
     \vspace{-3ex}
    \label{fig:battery_n_latency_combined}
\end{figure}

The larger pool is useful only if accessing it does not require
executing all of its members. Figure~\ref{fig:battery_n_latency_combined}
therefore measures the systems cost as more models participate in each
inference. The left plot reports battery drain per sample, and the right
plot reports inference latency under parallel and serial execution.
Battery consumption increases even with parallel execution, while
latency also grows as more models are executed. Because continuous
sensing repeats this computation window after window, these costs
accumulate throughout deployment.

Together, Figures~\ref{fig:motivation_acc_vs_n} and
\ref{fig:battery_n_latency_combined} motivate the operating point we
target: retain access to a larger pool of $N$ models, but execute only
$K \ll N$ members at each sensing window. The larger pool provides
alternatives that can improve the composition of a fixed $K$-member
ensemble, while limiting execution to $K$ members preserves the
computational advantage of selective inference.

\subsection{Short-Term Persistence Supports Active-Set Maintenance}
\label{sec:feasibility:persistence}

Executing only $K$ models still leaves a systems question: should the
selected $K$ members be reconsidered at every sensing window, or can
the current \textit{active set} be carried forward? Maintaining the
active set reduces model replacement only if per-model reliability
persists across nearby inputs. We therefore measure this property
directly from the base-model predictions, independently of
\system{}.

\begin{figure}[t]
    \centering
    \includegraphics[width=.8\linewidth]{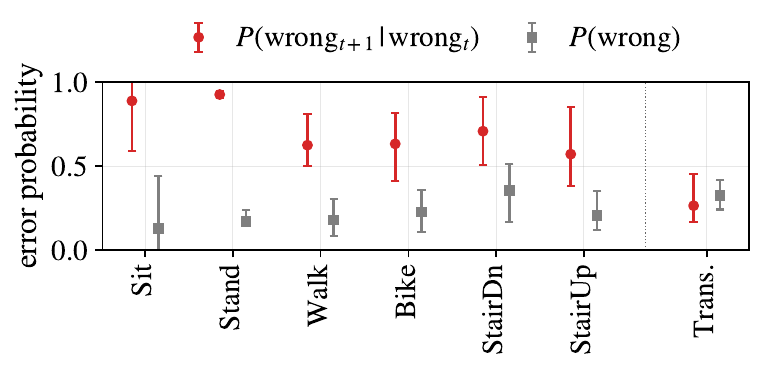}
    \vspace{-3ex}
    \caption{Per-class error persistence. Red and gray show the conditional next-window and overall error rates, respectively. The rightmost bucket denotes activity transitions. Markers and bars indicate the mean and min--max range across pool members.}
    \vspace{-3ex}
    \label{fig:temporal_active_set}
\end{figure}

Figure~\ref{fig:temporal_active_set} reports error persistence for each
activity. For every pool member, we compare its ordinary error
probability, $P(\mathrm{wrong})$, with the probability that it remains
wrong on the next window after being wrong on the current one,
$P(\mathrm{wrong}_{t+1}\mid\mathrm{wrong}_t)$. The vertical axis reports
these probabilities, markers show the mean across pool members, and
error bars show their min--max range. The rightmost bucket separately
reports adjacent windows in which the activity changes.

For consecutive windows within the same activity, the conditional error
probability is substantially higher than the ordinary error rate across
all six activities. A model that fails on the current window is therefore
much more likely to remain unreliable on the following window than its
average error rate would suggest. This effect becomes much weaker across
activity transitions, indicating that reliability tends to persist
within a sensing context and changes more strongly when that context
changes.

This persistence makes per-window reselection unnecessary. Rather than
reconstructing the entire $K$-member ensemble at every window, the
system can preserve the current active set and reconsider its composition
only when there is evidence that one of its members has become
unreliable. This directly reduces the model replacements required for
adaptive selection.

\subsection{Updating the Active Set Without Candidate Execution}
\label{sec:feasibility:update}

Maintaining the active set reduces adaptation to two decisions: whether
a currently active member should remain selected, and, if not, which
inactive member should replace it. Making both decisions by probing
candidate models would require additional forwards beyond the $K$
models already executed. We therefore examine whether the two decisions
can instead use information that is already available or can be
prepared before deployment.

\begin{figure}[t]
    \centering
    \includegraphics[width=\linewidth]{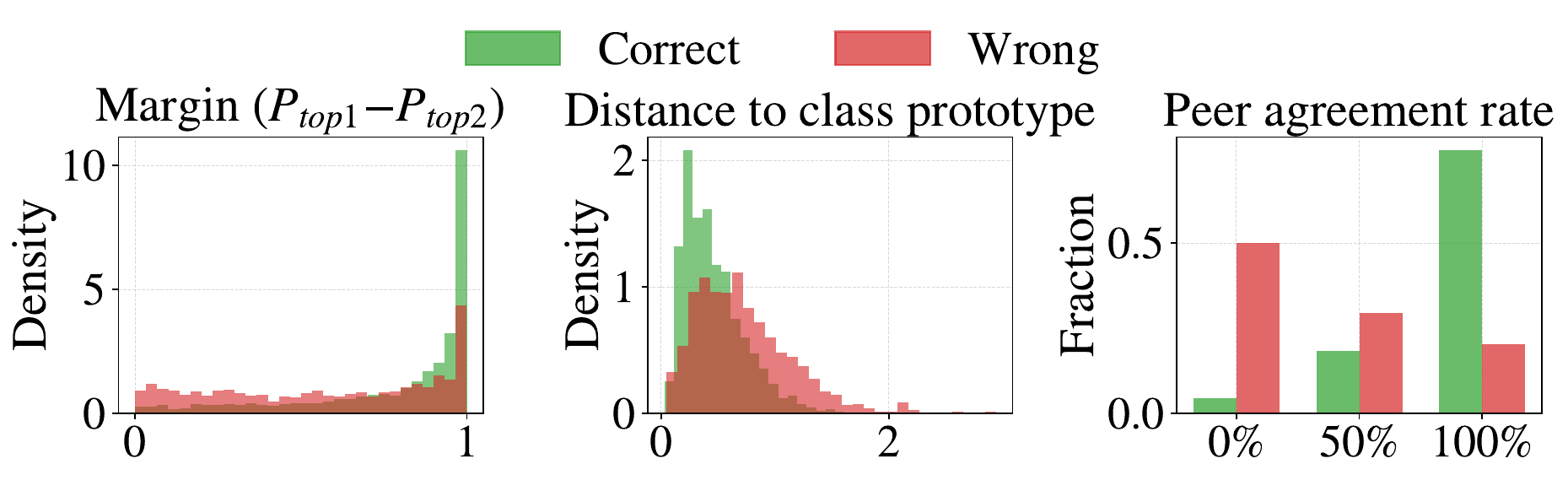}
    \vspace{-3ex}
    \caption{Distributions of runtime signals for correct and wrong
    predictions of an active member.}
     \vspace{-3ex}
    \label{fig:rejecter_runtime_signals}
\end{figure}

We first consider the decision to remove an active member. Because an
active model has already executed on the current input, its prediction
and intermediate representation are available at no additional
model-forward cost. Figure~\ref{fig:rejecter_runtime_signals} examines
whether these outputs contain useful evidence of correctness. The three
plots compare correct and wrong predictions using prediction margin,
Mahalanobis ratio, and agreement with the other active members.
Prediction margin reflects output confidence, the Mahalanobis ratio
measures how typical the current representation is for the predicted
class, and peer agreement captures support from the rest of the active
set. Greater separation between the correct and wrong distributions
indicates a more informative reliability signal.

The three signals expose complementary aspects of model reliability.
Although no single signal perfectly identifies errors, their
distributions differ between correct and wrong predictions. More
importantly, all of them are obtained from the $K$ models that have
already been executed. This suggests using rich \textit{post-execution}
evidence to determine whether each active member should remain selected,
without probing any inactive model.

Selecting a replacement presents the opposite situation. An inactive
model has not yet executed, so obtaining comparable input-specific
evidence would require an additional forward. We therefore examine
whether replacement quality contains structure that can be summarized
before deployment.

\begin{figure}[t]
    \centering
    \includegraphics[width=.99\linewidth]{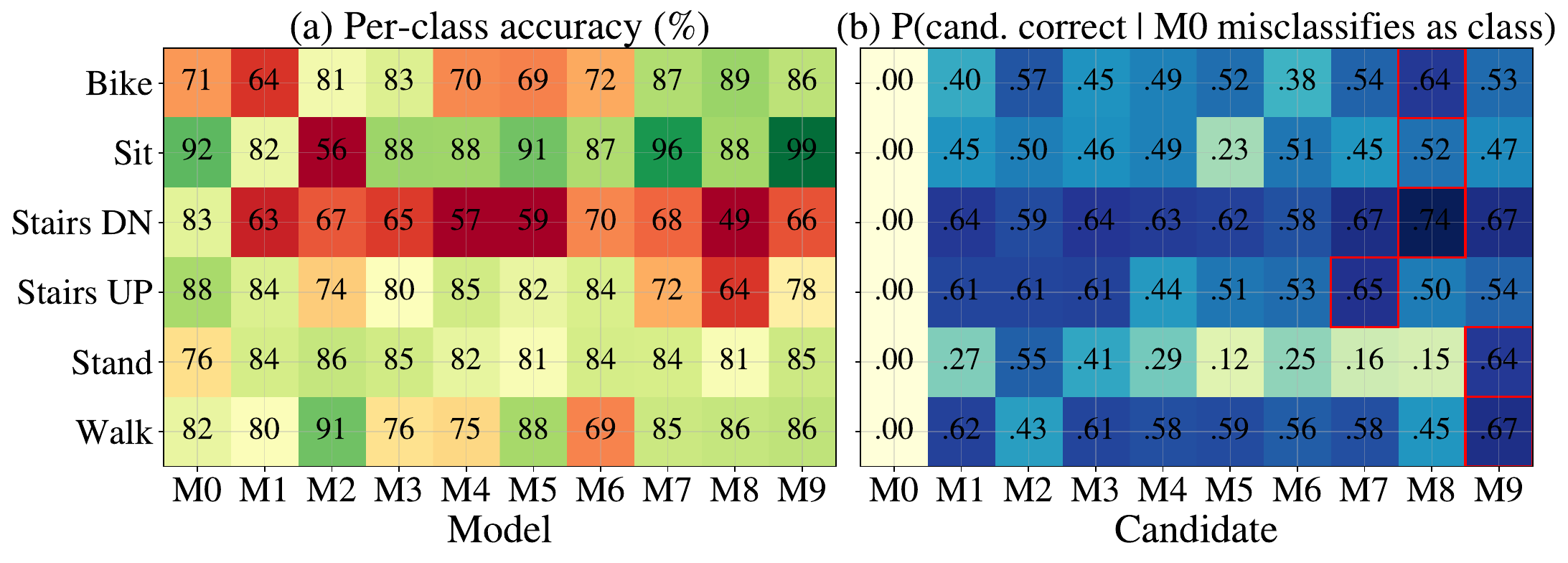}
    \vspace{-3ex}
    \caption{Class-conditional replacement evidence on HHAR. (a) Per-class accuracy of pool members. (b) Candidate correctness conditioned on each incorrect class prediction by a representative member. Red boxes indicate the best replacement.}
    \vspace{-3ex}
    \label{fig:router_evidence}
\end{figure}

Figure~\ref{fig:router_evidence} (a) reports the per-class accuracy of
the ten pool members. Each column represents a model and each row an
activity class. Despite sharing the same architecture and training
configuration, the models exhibit different class-level strengths.
Figure~\ref{fig:router_evidence} (b) examines this complementarity more
directly for a representative failed member. Rows correspond to the
class incorrectly predicted by that member, columns correspond to
candidate replacements, and each cell reports the probability that the
candidate predicts the correct label under that failure. Higher values
indicate a better replacement.

The strongest replacement changes across predicted classes, showing
that a single global ranking of inactive models is insufficient.
Instead, replacement quality depends on both which member failed and
what that member predicted. This structure can be summarized before
deployment as a class-conditional ranking indexed by the rejected member
and its predicted class. An inactive replacement can then be selected
without executing candidate models.

Together, these results reveal an asymmetry in the information available
for active and inactive models. Rich input-specific evidence is already
available after an active member executes, whereas obtaining the same
type of evidence from an inactive candidate would require another
forward. We therefore use post-execution evidence to identify an
unreliable active member and precomputed class-conditional
complementarity only after a replacement becomes necessary. Combined
with the temporal persistence above, this motivates a stateful
\textit{reject-then-route} strategy that preserves the active set by
default and updates only the members that need to be replaced.

\begin{figure*}[t]
    \centering
    \includegraphics[width=0.7\linewidth]{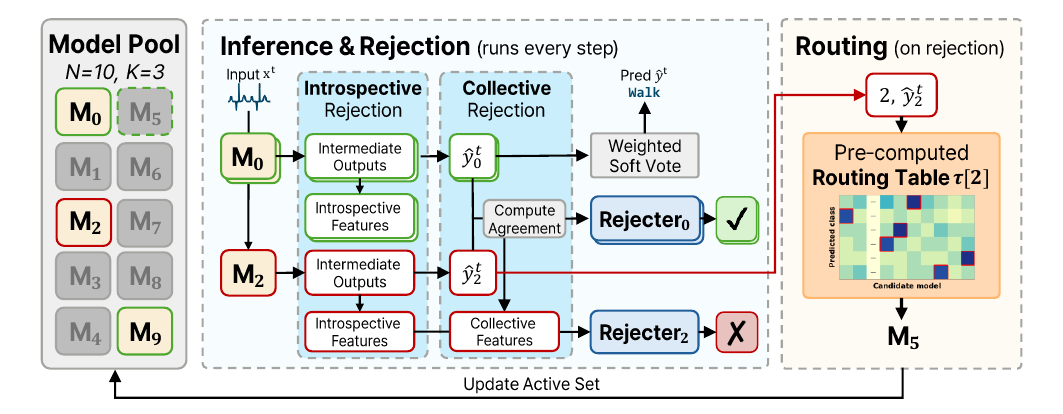}
    \caption{Overview of \system{}.} 
    \label{fig-overview}
\end{figure*}

\section{Design of \system{}}
\label{sec-design}

\system{} realizes selective ensemble inference by maintaining a small
set of models across consecutive sensing windows and updating only the
members that become unreliable. Let
$\mathcal{M}=\{f_1,\ldots,f_N\}$ denote the full model pool and
$\mathcal{A}_t\subset\mathcal{M}$ the \emph{active set} of $K$ models
executed at sensing window $t$, where $K\ll N$. Rather than repeatedly
selecting $K$ models from the entire pool, \system{} preserves
$\mathcal{A}_t$ by default and follows a \emph{reject-then-route}
procedure only when its composition needs to change. This design
addresses the two overheads identified in
Section~\ref{sec:feasibility}, unnecessary model replacement and
additional candidate-model inference.

\subsection{Overall Framework}
\label{sec-design-workflow}

Figure~\ref{fig-overview} illustrates the overall operation of
\system{}. The framework treats the active set as runtime state that
evolves with the sensing stream. At each window, the $K$ active models
first process the input in parallel. Their outputs are used both to
produce the current prediction and to determine whether the same models
should remain active afterward.

Before deployment, \system{} prepares two lightweight components for
this decision process. Each base model is associated with a
\emph{rejecter} that learns to distinguish its correct and incorrect
predictions using information available after inference. The training
data are also used to summarize which pool members tend to correct
particular failures of other members in a class-conditional
\emph{routing table}. The rejecter consequently handles an
input-specific decision about a model that has already executed. The
router handles a different decision about models that have not executed,
using model complementarity learned beforehand.

Runtime inference proceeds in four steps. The models in
$\mathcal{A}_t$ first execute on the incoming input. Their rejecters
then determine which predictions should remain trusted. The accepted
members produce the current prediction through margin-weighted soft
voting. If any member is rejected, the router fills only the
corresponding position using an inactive model selected without
candidate inference. Accepted members remain unchanged and the selected
replacement participates from the following window.

This separation is central to \system{}. Rich evidence is used where
it is already available after active-model inference. Selection among
inactive models instead relies on information that does not require
another forward pass. The following sections explain how these two
principles allow the active set to evolve without losing the efficiency
of $K$-model execution.

\subsection{Maintaining the Active Set Across Windows}
\label{sec-design-active}

The first design decision concerns when the selected ensemble should
change. An adaptive selector could reconsider all $K$ positions at
every sensing window. The short-term reliability persistence observed
in Section~\ref{sec:feasibility}, however, suggests that
this would repeatedly reconsider models whose usefulness has not yet
changed. Such updates can introduce model loading and memory-transfer
cost even when most of the current active set remains suitable.

\system{} therefore follows a conservative update policy. An active
member remains in the set as long as there is no evidence that it has
become unreliable. If every member in $\mathcal{A}_t$ remains accepted,
the active set is simply carried forward and
$\mathcal{A}_{t+1}=\mathcal{A}_t$. When one or more members are
rejected, only those positions are reconsidered. The accepted members
remain unchanged.

This is how \system{} exploits temporal continuity. The framework does
not assume that one ensemble remains optimal throughout an activity or
sensing context. It instead avoids discarding evidence supporting
members that continue to behave reliably. A local failure therefore
changes only the affected part of the active set rather than triggering
a new selection of all $K$ models.

A rejected member is excluded from the accepted vote for the current
window. Its position is filled at the end of the same window and the
replacement becomes active at window $t+1$. There is consequently no
multi-window rejected state. What persists across windows is the
portion of the active set that has not been rejected.

The accepted members produce the current prediction through
margin-weighted soft voting. Each softmax probability vector is
weighted by its \emph{probability margin}, defined as the difference
between the largest and second-largest class probabilities. If all
$K$ active members are rejected at the same window, \system{} falls
back to soft voting over the full current active set and leaves its
composition unchanged for the following window. This avoids rebuilding
the entire active set from an ambiguous all-reject event.

At the beginning of a sensing stream, no active set is available.
\system{} therefore performs a one-time \emph{auto-initialization}.
All $N$ pool members execute on the first input and are evaluated by
their corresponding rejecters. The $K$ models with the highest
acceptance confidence form the initial active set. Subsequent windows
execute only these $K$ active models unless a rejection causes one of
their positions to change.

\subsection{Post-Execution Member Rejection}
\label{sec-design-rejecter}

Once the active set is preserved by default, runtime selection becomes
a more focused problem. After an active model has already executed,
\system{} must determine whether there is enough evidence to stop
trusting its current prediction. This ordering is important because the
required $K$ forwards have already produced predictions and internal
representations. The rejection decision can therefore use rich
input-specific information without adding another base-model forward.

A single confidence value is not sufficient for this decision. An
incorrect model can still produce a confident output, and a model with
moderate confidence can still be correct. \system{} therefore examines
each prediction from several complementary perspectives. It considers
how strongly the model supports its own decision, whether its internal
representation supports that class, whether the other active members
provide consistent evidence, and whether the model has behaved stably
over recent windows.

The model's output provides the first source of evidence. Its softmax
probability vector preserves the complete distribution over classes,
while the \emph{probability margin} between the two largest
probabilities summarizes how clearly the model prefers its selected
class.

The internal representation provides a different view. For pooled
representation $h_k(x)$, \system{} computes a
\emph{Mahalanobis distance ratio} between the distance to the
predicted-class centroid and the distance to the nearest alternative
class centroid. The required class centroids and covariance statistics
are estimated from the training data. This signal captures whether the
current representation resembles the region in which the model
normally makes the corresponding prediction.

The other active models provide additional evidence about the same
input. The \emph{agreement rate} measures the fraction of active peers
that predict the same class as $f_k$. The \emph{dissenter margin}
captures the strongest probability margin among peers that disagree.
These signals do not treat majority agreement as ground truth. They
instead indicate whether the current prediction is supported or
strongly challenged by the rest of the active ensemble.

Finally, \system{} incorporates recent temporal behavior through a
\emph{flick} signal. It counts how often the predicted label of the
member changes over its five most recent windows. This provides a
short-term view of prediction stability that complements the
instantaneous evidence from the current input. Its inclusion follows
the temporal persistence observed in
Section~\ref{sec:feasibility}.

Each base model $f_k$ has its own lightweight rejecter $r_k$ that
combines these signals. We use separate rejecters because independently
trained ensemble members can develop different failure patterns even
when they share the same architecture and training procedure. The
rejecter therefore learns when a particular member becomes unreliable
rather than imposing one shared notion of uncertainty across the
entire pool.

Each rejecter is trained offline as a correctness detector for its
corresponding model. For a labeled sample $(x,y)$, the target is
\emph{reject} whenever $f_k(x)\neq y$ and \emph{accept} otherwise.
The target depends only on whether $f_k$ itself predicts correctly.

Optimization uses examples from the training split, while validation
examples are reserved for checkpoint selection. We use two stages
because the rejecter serves both an individual-model decision and an
ensemble-level objective. The first stage selects a checkpoint using
validation binary accuracy, encouraging the rejecter to distinguish
correct from incorrect predictions of its base model. The second stage
again trains on the training examples, but chooses the deployed
checkpoint using validation system accuracy measured over sampled
$K$-member ensembles. This second criterion reflects the purpose of
the rejecter within \system{}. A useful rejection decision should not
only identify an individual error, but should improve the prediction
produced by the active ensemble.

\subsection{Forward-Free Replacement Routing}
\label{sec-design-router}

Rejecting a member determines where the active set should change. It
still leaves the problem of choosing which inactive model should take
that position. This decision has a different information constraint
from rejection. The rejected member has already executed on the
current input, whereas an inactive candidate has not. Executing
candidates to obtain their predictions or feature representations
would provide stronger input-specific evidence, but every such probe
would add a model forward and reduce the computational saving of
$K$-model execution.

\system{} instead uses model complementarity learned before deployment.
The motivating analysis in Section~\ref{sec:feasibility} shows
that useful replacements exhibit class-conditional structure. A model
that complements one failure is not necessarily the best replacement
for another. The appropriate replacement depends on both which member
failed and the class predicted by that member.

For every rejected model $f_k$, predicted class $c$, and candidate
model $f_j$, \system{} estimates
$\tau_{k,c,j}=\Pr\!\left(f_j(x)=y\mid f_k(x)=c\right),$ where $y\neq c$
using the training data. The value $\tau_{k,c,j}$ measures how often
$f_j$ correctly handles examples on which $f_k$ incorrectly predicts
class $c$. These values form a class-conditional routing table that
summarizes pairwise model complementarity before deployment.

At runtime, when $f_k$ is rejected after predicting class $c$, the
router retrieves $\tau_{k,c,\cdot}$ and ranks the inactive members by
their corresponding scores. The highest-ranked candidate fills the
rejected position. The decision requires neither a candidate prediction
nor a candidate feature representation, so no additional base-model
forward is introduced.

The table-based router follows directly from the information asymmetry
between active and inactive models. Rejection can use current
input-specific evidence because that evidence has already been produced
by required inference. Routing intentionally avoids demanding the same
evidence from inactive candidates because obtaining it would require
the additional computation that selective execution is designed to
avoid. Offline complementarity is therefore used in place of runtime
candidate probing.

When multiple members are rejected in the same window, each inactive
candidate is evaluated against all rejected positions. Its maximum
class-conditional replacement score across those positions is used for
ranking. The accepted members remain in place and the highest-ranked
candidates fill the vacated positions together in a single active-set
update.

Together, these components implement the operating point established
in Section~\ref{sec:feasibility}. Active-set maintenance avoids
unnecessary model replacement, post-execution rejection extracts
reliability evidence from computation that has already been performed,
and forward-free routing selects alternatives without probing inactive
models. Regular inference can therefore adapt the composition of the
ensemble while remaining close to the computational cost of executing
only $K$ models.


\begin{table*}[t]
\centering\small
\begin{adjustbox}{width=0.85\linewidth}
\begin{tabular}{@{}l cccc cccc cccc cccc@{}}
\toprule
 & \multicolumn{4}{c}{HHAR}
 & \multicolumn{4}{c}{UCIHAR}
 & \multicolumn{4}{c}{WISDM}
 & \multicolumn{4}{c}{MOTIONSENSE} \\
\cmidrule(lr){2-5}
\cmidrule(lr){6-9}
\cmidrule(lr){10-13}
\cmidrule(lr){14-17}

 & Trf & CNN & GRU & MLP
 & Trf & CNN & GRU & MLP
 & Trf & CNN & GRU & MLP
 & Trf & CNN & GRU & MLP \\

\midrule

\textit{Oracle}
& \textit{95.3} & \textit{95.7} & \textit{96.3} & \textit{94.0}
& \textit{93.9} & \textit{95.6} & \textit{94.8} & \textit{91.9}
& \textit{95.2} & \textit{97.9} & \textit{97.4} & \textit{87.4}
& \textit{96.6} & \textit{95.1} & \textit{93.8} & \textit{93.3} \\

\midrule

Avg Single
& 75.5 & 87.8 & 86.9 & 84.7
& 84.8 & 91.6 & 88.2 & 87.4
& 83.4 & 91.2 & 82.8 & 81.2
& 84.1 & 88.9 & 81.9 & 87.8 \\

Best Single
& 79.1 & 91.5 & 88.7 & 88.4
& 86.9 & 92.7 & 89.5 & 88.5
& 87.4 & 93.5 & 85.2 & 83.6
& 88.5 & 92.2 & 86.2 & 89.3 \\

\midrule

Static $K{=}3$
& 87.8 & 91.5 & 89.8 & 86.9
& 85.1 & 91.2 & 90.3 & 87.5
& 85.2 & 92.0 & 85.6 & 81.6
& 87.2 & 89.5 & 85.3 & 88.4 \\

$N{=}10$ Soft
& \textbf{89.9} & \textbf{93.0} & 91.2 & 87.6
& 87.8 & 92.6 & \textbf{91.0} & \textbf{88.0}
& 85.3 & 93.8 & 86.7 & \textbf{81.9}
& 89.0 & 90.4 & 86.5 & 88.4 \\

\midrule

\system{} ($K{=}3$)
& \cellcolor{green!12}89.1
& \cellcolor{green!12}92.6
& \cellcolor{green!12}\textbf{91.4}
& \cellcolor{green!12}\textbf{88.3}

& \cellcolor{green!12}\textbf{88.3}
& \cellcolor{green!12}\textbf{92.8}
& 90.2
& \cellcolor{green!12}87.8

& \cellcolor{green!12}\textbf{86.7}
& \cellcolor{green!12}\textbf{94.8}
& \cellcolor{green!12}\textbf{88.3}
& 81.6

& \cellcolor{green!12}\textbf{90.3}
& \cellcolor{green!12}\textbf{91.9}
& \cellcolor{green!12}\textbf{86.8}
& \cellcolor{green!12}\textbf{89.1} \\

\bottomrule
\end{tabular}
\end{adjustbox}

\caption{Classification accuracy (\%) across four HAR datasets
($N{=}10$, $K{=}3$).
Green marks cases where \system{} improves on Static $K{=}3$ at the same
forward budget.
Bold marks the better result between \system{} and full $N{=}10$ soft voting.}
\vspace{-5ex}
\label{tab:e1_cross_dataset}
\end{table*}

\section{Evaluation}
\label{sec:eval}

\begin{figure*}[t]
    \centering
    \includegraphics[width=.7\linewidth]{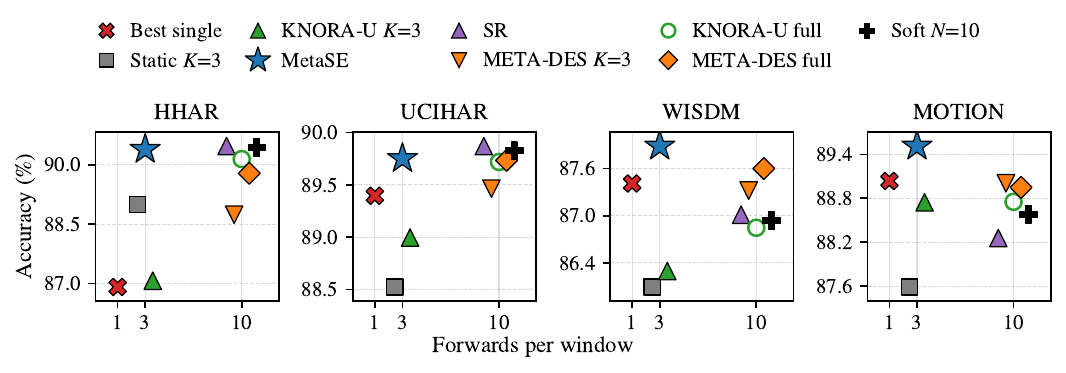}
    \vspace{-4ex}
    \caption{Accuracy and base-model forward cost across four HAR datasets.
    Each point reports mean accuracy across four model architectures.
    The horizontal position reports the number of base-model forwards
    required per regular sensing window.}
    \label{fig:eval-acc-forwards}
\end{figure*}

The evaluation follows the operating point established in the preceding
sections.
We first examine how much prediction quality can be obtained while regular
inference is limited to three members, and compare this operating point with
both conventional ensembles and existing dynamic selection methods.
We then examine where selective execution provides its gains and how the
rejecter shapes the active ensemble.
Finally, we study the temporal evolution of the selected members and measure
the resulting cost on an embedded platform.


\subsection{Evaluation Setup}
\label{sec:eval-setup}

\noindent{}\textbf{$\bullet$ Datasets.}
We evaluate \system{} on four public HAR datasets spanning different users,
devices, sensing configurations, and activity sets.
HHAR~\cite{stisen2015smart} contains smartphone accelerometer and gyroscope
measurements from 9 users performing 6 activities.
UCI-HAR~\cite{anguita2013public} contains smartphone inertial measurements
from 30 subjects performing 6 activities.
WISDM~\cite{weiss2019wisdm} contains accelerometer measurements from
36 users across 6 activities.
MOTIONSENSE~\cite{malekzadeh2019mobile} contains iPhone accelerometer and
gyroscope measurements from 24 subjects performing 6 activities.
All evaluations are subject-disjoint.
Held-out users account for roughly 20--30\% of each dataset, with UCI-HAR
retaining its canonical nine-subject test split.

\noindent{}\textbf{$\bullet$ Model pools.}
We evaluate Transformer (Trf), CNN, GRU, and MLP pools to cover model families
with different temporal and representational structures.
Each pool contains $N{=}10$ independently trained instances of the same
architecture with different random initialization and data shuffling.
Unless otherwise stated, \system{} maintains $K{=}3$ active members.

\noindent{}\textbf{$\bullet$ Ensemble references.}
Static $K{=}3$ starts from the same initial active set as \system{} and keeps
those members fixed throughout the stream, providing the primary comparison
under the same three-model forward budget.
Full $N{=}10$ soft voting executes the complete pool and represents
conventional exhaustive ensemble inference.
We additionally report average single-model accuracy, the best individual
test member, and Oracle, which indicates whether at least one member in the
full pool predicts each input correctly.

\noindent{}\textbf{$\bullet$ Adaptive selection baselines.}
We compare against KNORA-U~\cite{ko2008dynamic} and META-DES~\cite{cruz2015meta} as representative
Dynamic Ensemble Selection methods, together with Softmax Response
(SR)~\cite{geifman2017selective} as a confidence-based selective-prediction
baseline.
KNORA-U estimates classifier competence from a local region around the
current query and selects competent members for prediction.
META-DES learns a meta-classifier that estimates the competence of each
member for the current query.
SR instead filters member predictions according to their maximum softmax
probability using thresholds calibrated on the validation set.
For all methods, we report both prediction accuracy and the number of
base-model forwards required to obtain the information used for selection.

\noindent{}\textbf{$\bullet$ HAR-specific ensemble baselines.}
We additionally compare against Ens-CLRF~\cite{mekruksavanich2024ensemble}
and FilterAct~\cite{huang2022deep}, selected from prior HAR methods because
both explicitly exploit multiple models or ensemble-style training.
Ens-CLRF combines multiple deep predictors for sensor-based activity
recognition, while FilterAct transfers information across multiple CNNs
during training and retains a single network for inference.
These methods provide task-specific ensemble comparisons alongside the
generic adaptive-selection baselines above.



\subsection{Overall Performance}
\label{sec:eval-overall}

We first examine the operating point targeted by \system{}.
A larger model pool provides useful alternatives, but exploiting this
diversity is attractive only if regular inference can remain limited to a
small active set.
We therefore compare \system{} with Static $K{=}3$, which keeps the same
initial three members throughout the stream, and full $N{=}10$ soft voting,
which executes every available member.
Test windows retain their temporal order.
After a one-time full-pool initialization, \system{} executes only the three
active members, replaces rejected positions after the current prediction,
and uses the selected replacements from the following window.

Table~\ref{tab:e1_cross_dataset} shows that adapting the three active
positions consistently improves on keeping them fixed.
\system{} outperforms Static $K{=}3$ in 14 of the 16 dataset and
architecture configurations, despite starting from the same active set and
using the same regular forward budget.
Averaged across the configurations, this adaptive three-member ensemble
also gains more over Static $K{=}3$ than simply expanding the vote to all
ten members.
These results show that the benefit comes not only from having access to a
larger pool, but also from adapting which part of that pool participates in
prediction.
\system{} further exceeds the retrospective Best Single reference in
10 of 16 configurations.
This result shows that the gain cannot generally be reproduced by choosing
one strong member and keeping it fixed.

Full-pool voting provides a complementary view of the same result.
Although $N{=}10$ soft voting uses more than three times as many model
forwards, \system{} is more accurate in 11 of the 16 configurations, and
its largest deficit in the remaining cases is only 0.8 percentage points.
Full soft voting itself also falls below the best individual member in
several configurations.
These results show that executing and aggregating more members does not
necessarily translate the available model diversity into a better
prediction, and that selective execution can retain access to that diversity
without involving every member at every sensing window.

Figure~\ref{fig:eval-acc-forwards} broadens the comparison to static
ensembles, confidence-based filtering, and representative Dynamic Ensemble
Selection methods.
Each point averages the four architecture-specific results within a dataset,
and its horizontal position gives the number of base-model forwards required
per regular sensing window.
This comparison separates the size of the final selected ensemble from the
cost of obtaining the evidence used to select it.
KNORA-U with $K{=}3$ operates at the same three-forward budget as
\system{}, whereas META-DES and Softmax Response require outputs from the
complete pool in our evaluation.

At the same three-forward budget, \system{} outperforms KNORA-U in 15 of
the 16 dataset and architecture configurations.
The advantage also extends beyond methods operating under the same budget.
After averaging across the four architectures, \system{} exceeds the
full-pool KNORA-U and META-DES variants on all four datasets despite
executing only three members.
Full ten-model soft voting provides the strongest exhaustive reference.
Despite executing only three members, \system{} remains within
0.1 percentage points of it on HHAR and UCIHAR and achieves higher average
accuracy on WISDM and MOTIONSENSE.

These results highlight an important difference in how selection evidence is
obtained.
Pool-wide methods can make decisions from richer current information, but
obtaining that information requires executing the models being considered.
\system{} instead uses richer post-execution evidence from the three members
that already run and revisits an active position only when its current member
is rejected.
As Figure~\ref{fig:eval-acc-forwards} shows, this allows \system{} to combine
a three-forward execution budget with accuracy comparable to or higher than
methods that inspect the complete pool.

\begin{figure}[t]
    \centering
    \includegraphics[width=.9\linewidth]{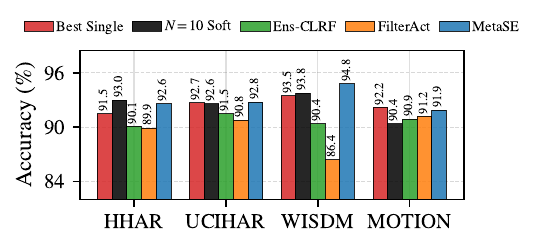}
    \vspace{-3ex}
    \caption{Accuracy comparison with HAR-specific ensemble methods using
    the CNN configuration.
    The comparison includes the best individual CNN, full $N{=}10$ soft
    voting, Ens-CLRF, FilterAct, and \system{} with $K{=}3$ active members.}
    \vspace{-3ex}
    \label{fig:eval-external}
\end{figure}

At the same three-forward budget, Figure~\ref{fig:eval-acc-forwards} shows
that \system{} outperforms KNORA-U in 15 of the 16 configurations, with
MOTIONSENSE/GRU as the only exception.
Both methods draw from the same pool and execute the same number of models,
so the difference reflects how that limited budget is used.
\system{} preserves members whose runtime evidence remains useful and
reconsiders individual positions after a detected failure, rather than
forming each prediction from a fresh pre-execution selection.
The result shows that adaptive selection under a small budget depends not
only on which models are chosen, but also on when existing choices should be
revisited.

The ten-forward methods expose the complementary tradeoff in
Figure~\ref{fig:eval-acc-forwards}.
Once the full set of outputs is available, Softmax Response reaches accuracy
close to \system{} on HHAR and UCIHAR, and the other full-pool methods occupy
a similar accuracy range on several datasets.
Obtaining this richer pool-wide evidence, however, requires executing all ten
models.
\system{} instead obtains richer post-execution evidence from the members
that are already active and uses lightweight information when an inactive
replacement is needed.
These results show that \system{} can approach the accuracy enabled by
pool-wide runtime evidence without paying the forward cost required to
obtain that evidence from every member.

The comparisons above use generic model-selection approaches over the same
underlying pools.
We next compare with Ens-CLRF and FilterAct, which are selected from prior
HAR methods because both explicitly exploit multiple models or
ensemble-style training.

\begin{figure}[t]
    \centering
    \includegraphics[width=.7\linewidth]{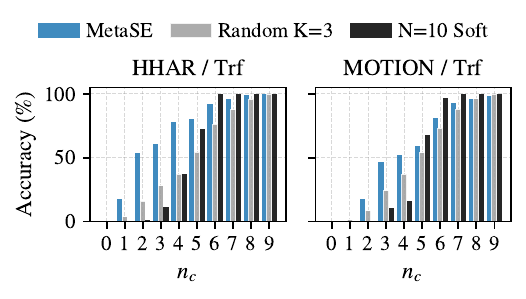}
    \vspace{-3ex}
    \caption{Streaming accuracy across $n_c$, the number of correct members
    in the full pool, on HHAR and MOTIONSENSE using Transformer pools.
    Numbers above the bars report the number of test windows in each group.}
    \vspace{-3ex}
    \label{fig:eval-nc-accuracy}
\end{figure}

\begin{figure*}[t]
    \centering
    \includegraphics[width=.9\linewidth]{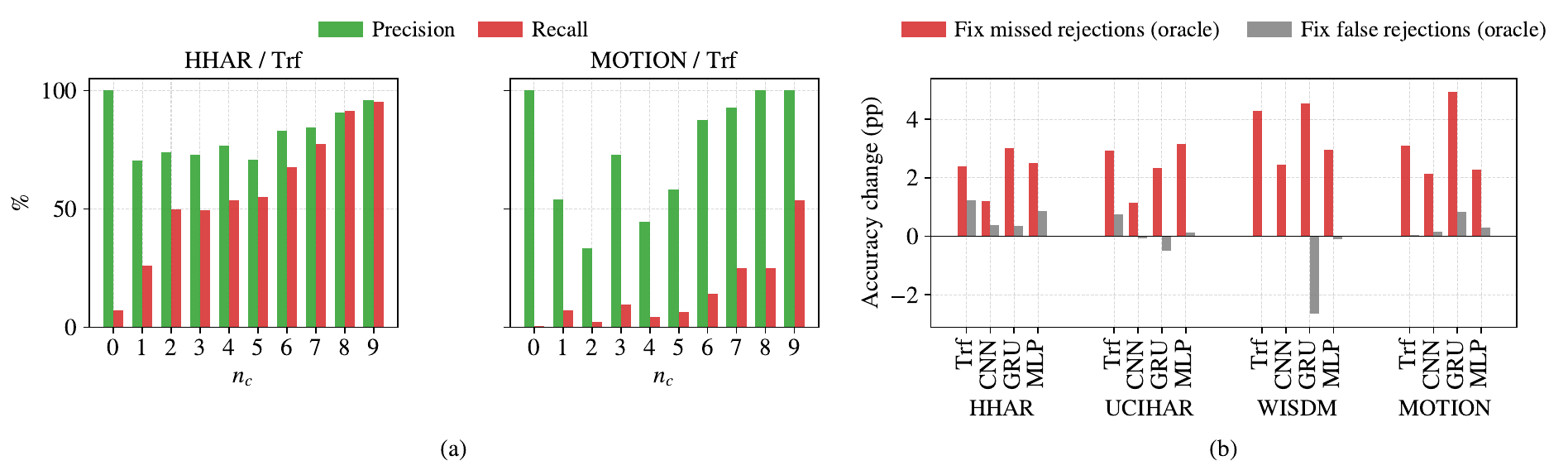}
    \caption{Rejecter analysis.
    (a) Rejection precision and recall across $n_c$ on HHAR and MOTIONSENSE.
    (b) End-to-end accuracy change when missed or false rejection decisions
    are corrected using oracle labels while the remaining streaming pipeline
    is unchanged.}
    \vspace{-2ex}
    \label{fig:eval-rejecter-analysis}
\end{figure*}

Figure~\ref{fig:eval-external} shows that \system{} remains competitive
with ensemble-based methods developed specifically for HAR.
It outperforms both Ens-CLRF and FilterAct on all four datasets.
Considering all methods in the figure, including the retrospective Best
Single reference and full $N{=}10$ soft voting, \system{} achieves the
highest accuracy on UCIHAR and WISDM and remains within 0.4 percentage
points of the best result on HHAR and MOTIONSENSE.
These results show that restricting regular execution to three members does
not require sacrificing the accuracy achieved by stronger task-specific or
full-ensemble alternatives.

The WISDM result further shows that this advantage is not explained only by
the quality of the underlying pool.
The best individual CNN already performs strongly at 93.5\%, but
\system{} improves it to 94.8\% and also exceeds full $N{=}10$ soft voting.
This result shows that an active ensemble can obtain value from model
diversity that neither a strong fixed member nor exhaustive aggregation
consistently captures.

\subsection{Selection Gain across Pool Correctness}
\label{sec:eval-nc}

Overall accuracy combines inputs for which the model pool offers very
different amounts of useful choice.
To understand where member selection actually contributes, we group each
input by $n_c$, the number of members in the complete pool that classify it
correctly.
Figure~\ref{fig:eval-nc-accuracy} reports this decomposition on the HHAR and
MOTIONSENSE Transformer pools.

Figure~\ref{fig:eval-nc-accuracy} shows that the aggregate gain does not come
mainly from inputs where nearly the entire pool fails.
Windows with $n_c{\leq}4$ account for only about 10--12\% of the analyzed
streams, and \system{} produces a net loss in the $n_c{\leq}2$ region on
HHAR.
These results show that large improvements on a small number of very
difficult inputs contribute relatively little to the overall accuracy gain.

Most of the gain instead comes from the contested middle of the pool.
Windows with $n_c{=}5$--$7$ account for 79\% of the improvement over
Static $K{=}3$ on HHAR and 59\% on MOTIONSENSE.
Several correct alternatives exist in this region, but the identity of the
three active members still affects the outcome.
At $n_c{\geq}8$, by contrast, most reasonable subsets already contain
sufficient correct support and selective composition contributes relatively
little.
Together, these results show that selective execution is most useful when
the pool contains multiple viable alternatives but the choice among them
still matters.

\subsection{Rejecter Behavior and Error Cost}
\label{sec:eval-rejecter}

The preceding analysis shows that member identity matters most when the pool
contains several plausible alternatives.
We therefore examine how reliably the rejecter identifies failing active
members and, more importantly, how its errors affect the complete streaming
system.
Figure~\ref{fig:eval-rejecter-analysis} (a) reports rejection precision and
recall across $n_c$.
For Figure~\ref{fig:eval-rejecter-analysis} (b), we replay each stream while
correcting only one type of rejecter error with oracle information, either
missed or false rejections, and leave the remaining decisions unchanged.
This intervention isolates the downstream cost of the two error types.

Figure~\ref{fig:eval-rejecter-analysis} (a) shows relatively high rejection
precision across much of the analyzed range, while recall varies more
strongly.
The system-level intervention in Figure~\ref{fig:eval-rejecter-analysis} (b)
shows that these errors have very different consequences.
Correcting missed rejections increases end-to-end accuracy by
2.84 percentage points on average across the 16 configurations, whereas
correcting false rejections changes accuracy by only 0.10 points.
These results show that the larger remaining opportunity lies in identifying
unreliable members that currently remain active.

Figure~\ref{fig:eval-rejecter-analysis} (b) also shows that member correctness
does not map directly to ensemble utility.
Correcting false rejections slightly reduces final accuracy in several
configurations because preserving one member can change both the current
vote and the subsequent active-set trajectory.
These results show that rejection decisions should be evaluated through
their effect on the ensemble rather than only through the correctness of
individual member-level decisions.



\subsection{Active-Set Update Dynamics}
\label{app:update-dynamics}

\begin{figure}[t]
    \centering
    \includegraphics[width=.8\linewidth]{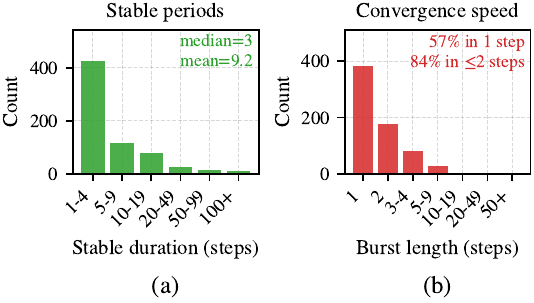}
    \vspace{-3ex}
    \caption{Temporal dynamics of active-set updates on HHAR/Transformer.
    (a) Distribution of consecutive windows without a replacement.
    (b) Distribution of consecutive replacement-burst durations.}
    \vspace{-3ex}
    \label{fig:app-update-dynamics}
\end{figure}

We further analyze how \system{} changes its active set over a continuous
sensing stream.
This analysis focuses on the temporal pattern of replacement rather than the
accuracy of the routing rule itself.
Using the HHAR/Transformer stream, we measure unchanged intervals as
consecutive windows without a replacement and replacement bursts as
consecutive windows in which one or more active positions are replaced.

Figure~\ref{fig:app-update-dynamics}(a) shows that unchanged intervals span
a broad range.
The median interval is three windows and the mean is 9.2, with some
selections remaining unchanged for substantially longer periods.
Figure~\ref{fig:app-update-dynamics}(b) shows that replacement activity is
usually short-lived, with 57\% of bursts ending after one window and 84\%
within two.
Thus, maintaining the active set does not imply keeping one fixed ensemble.
Selections can persist when their members remain useful, while subsequent
changes are typically confined to short update periods.

\begin{figure}[t]
    \centering

    \begin{subfigure}{0.62\linewidth}
        \centering
        \includegraphics[width=\linewidth]{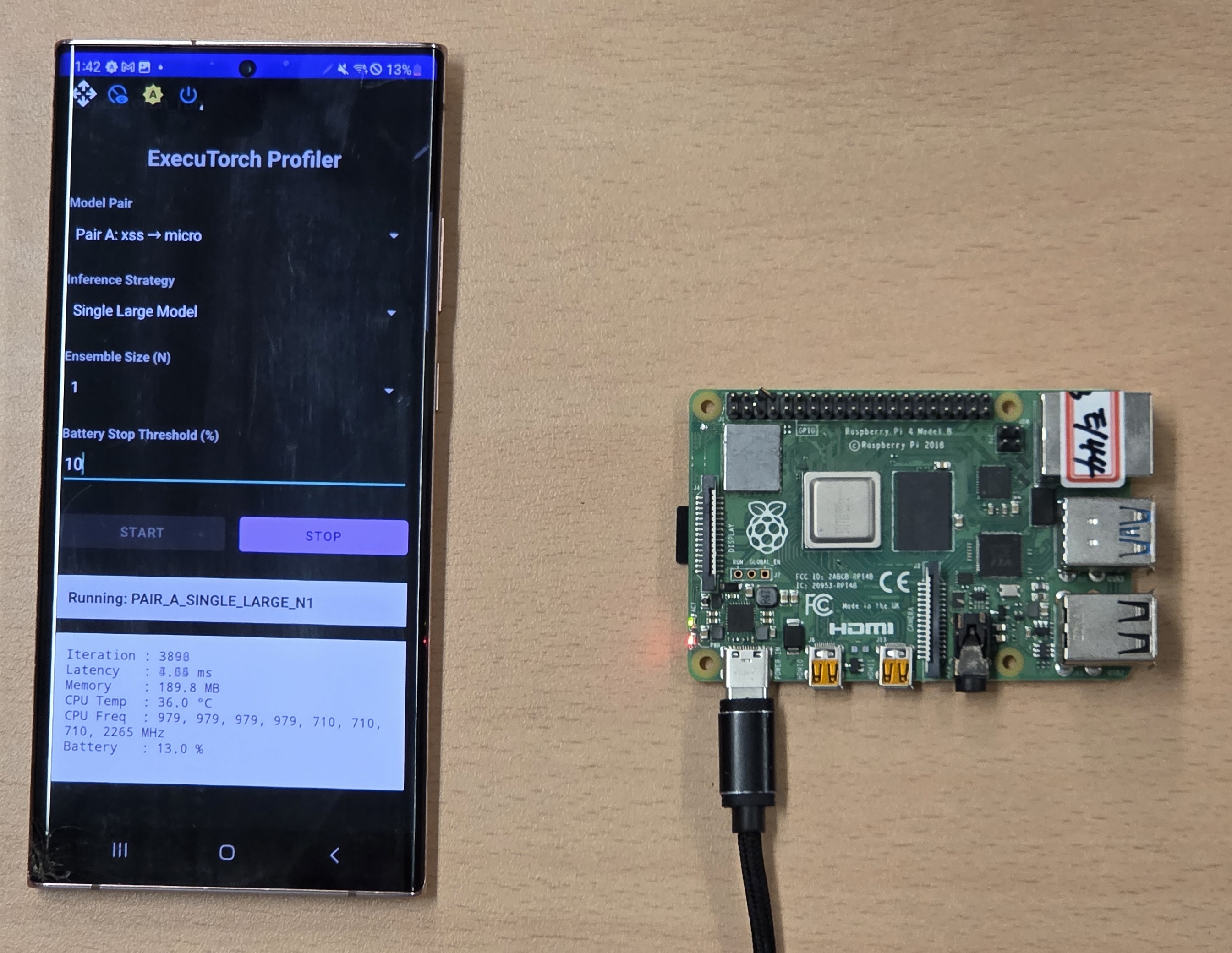}
        \caption{}
        \label{fig:testbed}
    \end{subfigure}

    \vspace{1ex}

    \begin{subfigure}{0.98\linewidth}
        \centering
        \includegraphics[width=\linewidth]{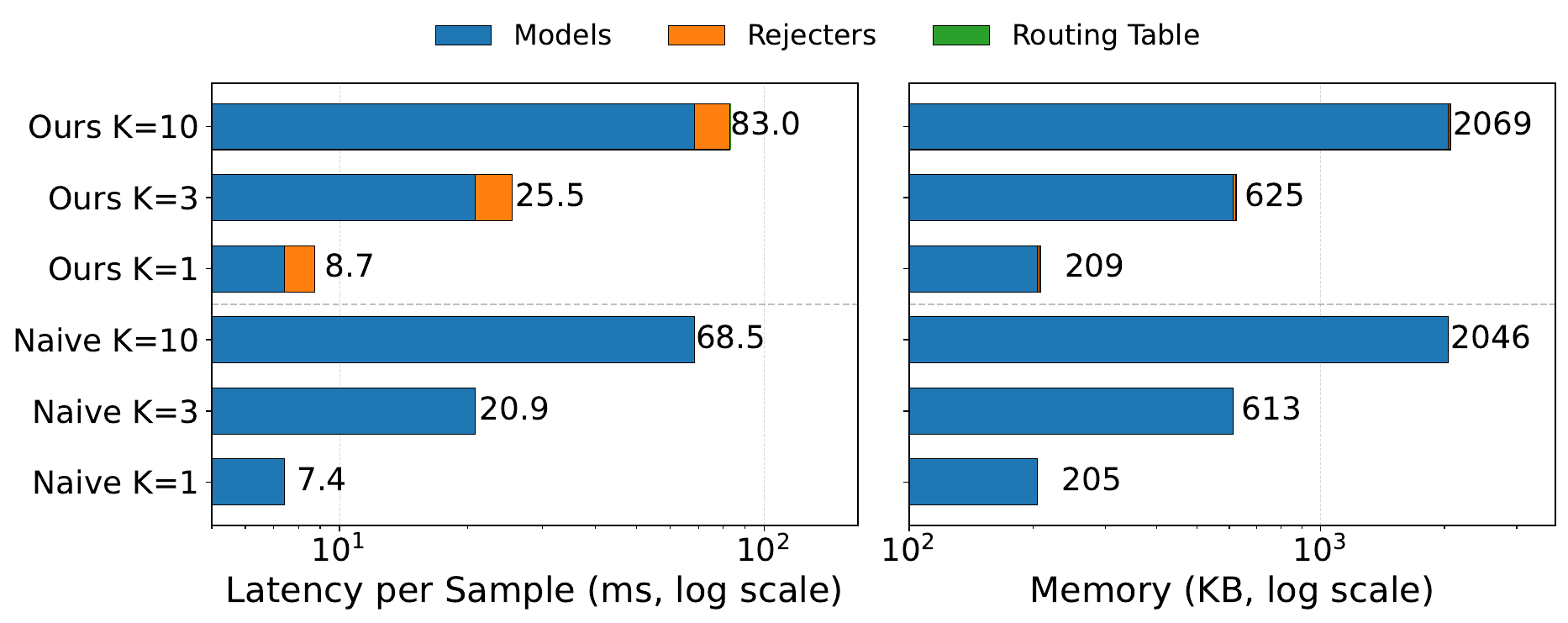}
        \caption{}
        \label{fig:device_overhead}
    \end{subfigure}
    \vspace{-2ex}
    \caption{On-device evaluation on Raspberry Pi 4B. 
    (a) Experimental testbed. 
    (b) Inference latency and resident memory across active-set sizes.}
    \label{fig:eval-ondevice}
    \vspace{-3ex}
\end{figure}

\subsection{On-Device Efficiency}
\label{sec:eval-ondevice}

The preceding experiments show that \system{} can adapt a three-member active
set without executing the complete model pool during regular inference.
We next examine whether the additional rejection and routing logic preserves
this computational advantage on a real embedded device.
As shown in Figure~\ref{fig:eval-ondevice} (a), we deploy \system{} on a
Raspberry Pi 4B and compare it with plain execution of one, three, and ten
base models.
The same-$K$ comparison captures the overhead introduced by adaptive
selection, while the $N{=}10$ configuration represents exhaustive
full-pool inference.

We measure sequential CPU inference latency and resident memory.
Each active member uses a 450-parameter rejecter, and the routing table
contains 600 floating-point entries.
A rejecter call takes approximately 0.02\,ms and a routing-table lookup
approximately 0.001\,ms; the end-to-end measurements below include the
complete reliability-management path.

Figure~\ref{fig:eval-ondevice} (b) shows that this additional logic keeps
the overall execution cost close to that of the underlying three-model
ensemble.
Plain $K{=}3$ inference requires 20.9\,ms and 613\,KB of resident memory,
compared with 25.5\,ms and 625\,KB for \system{}.
The difference becomes much larger relative to full ten-model inference,
which requires 68.5\,ms and 2,046\,KB.
Thus, even after accounting for rejection and routing, \system{} is
2.7$\times$ faster and uses 69\% less memory than full-ensemble inference.
These measurements confirm that adaptive active-set management preserves
most of the systems benefit of selective execution on the embedded device.
\section{Discussion}
\label{sec:discussion}

\vspace{0.5ex}
\noindent{}\textbf{Reliability under changing deployment conditions.}
\system{} relies on reliability signals learned before deployment, which may become less informative under distribution shifts.
Our analysis also suggests that missed rejections remain a major source of error.
Improving robustness to such changes without requiring pool-wide inference is an important direction for future work.

\vspace{0.5ex}
\noindent{}\textbf{Limited information for inactive members.}
Because \system{} does not execute inactive candidates, replacement decisions rely on lightweight historical information rather than current-input evidence.
This keeps inference cost low but may miss the best replacement for a given window.
Future designs could selectively probe inactive members when the expected benefit outweighs the additional cost.

\section{Conclusion}
\label{sec:conclusion}

We present \system{}, an active ensemble framework that maintains a small subset of models from a larger pool for continuous mobile sensing.
By preserving reliable members across nearby windows and replacing them only when needed, \system{} achieves accuracy comparable to or better than more expensive selection approaches while executing only three models during regular inference.
On a Raspberry Pi 4B, it is 2.7$\times$ faster and uses 69\% less memory than full ten-model inference.



\clearpage
\balance

\bibliographystyle{ACM-Reference-Format}
\bibliography{reference}

\appendix

\twocolumn[{%
\begin{minipage}{\textwidth}
    \centering
    \includegraphics[width=\textwidth]
    {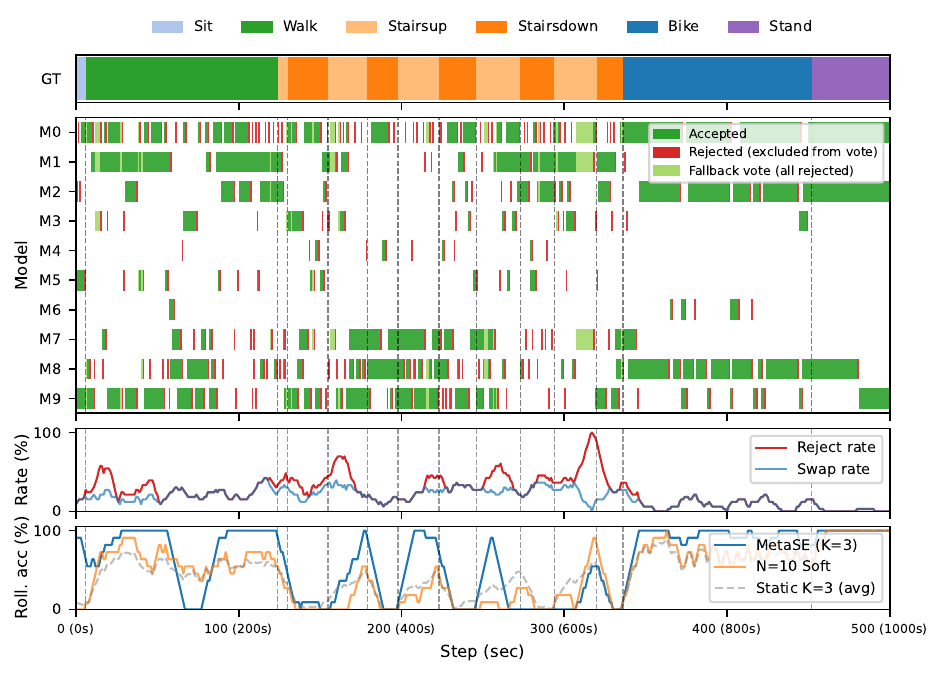}

    \captionof{figure}{Example streaming trace on HHAR/Transformer.
    From top to bottom, the panels show the ground-truth activity, accepted
    and rejected members, rolling rejection and replacement rates, and
    rolling accuracy of \system{}, full soft voting, and the fixed
    three-model reference.}
    \label{fig:app-streaming}
    \vspace{2ex}
\end{minipage}
}]

\section{Example Streaming Behavior}
\label{app:streaming-example}

To illustrate how member-level decisions appear within an actual sensing
sequence, Figure~\ref{fig:app-streaming} shows a 500-window segment of the
HHAR/Transformer stream.
This example is intended as a qualitative view of the runtime behavior,
complementing the aggregate update statistics above.

Figure~\ref{fig:app-streaming} shows that periods of rejection and
replacement are interspersed with intervals in which the current active
members remain unchanged.
The changes do not follow a fixed mapping from activity labels to model
subsets.
Updates can occur within an activity segment, and selected members can also
remain active across different portions of the stream.
The active set instead changes according to the observed reliability of its
current members.

The trace also illustrates the local nature of an update.
When one member is rejected, the unaffected members continue to participate
rather than forcing the complete active set to be selected again.
This produces localized changes while allowing previous selection decisions
to be reused when they remain valid.


\end{document}